\documentclass{llncs}
\usepackage{amsmath,amssymb,amsbsy}

\let\epsilon\varepsilon
\let\phi\varphi
\def\X{\{0,1\}^*}
\def\Y{\{0,1\}}

\def\N{\mathbb N}
\def\R{\mathbb R}

\def\C{\mathcal C}
\def\ep{0.05}

\begin{document}
\title{On computability of pattern recognition problems}
\author{Daniil Ryabko }
\institute{IDSIA, Galleria 2, CH-6928 Manno-Lugano, Switzerland\footnote{This work was supported by SNF grant 200020-107616}\\
\email{daniil@ryabko.net}}
\maketitle

\begin{abstract}
In  statistical setting of the pattern recognition problem the
number of examples required to approximate an unknown labelling
function is linear in the VC dimension of the target learning
class. In this work we consider the question whether such bounds
exist if we restrict our attention to computable pattern recognition methods,
assuming that the unknown labelling function is also computable.
We find that in this case the number of examples required for a
computable method to approximate the labelling function not only
is not linear, but grows  faster
 (in the VC dimension of the class) than any
computable function.  No time or space constraints are put on the
predictors or target functions; the only resource we consider is the training
examples.

The task of pattern recognition is considered in conjunction with another learning
problem~--- data compression. An impossibility result for the task of data compression
allows us to estimate the sample complexity for pattern recognition.
\end{abstract}

\section{Introduction}\label{sec:intro}
The task of pattern recognition consists in
  predicting an unknown label of some observation (or object). For instance,
the  object can be an image of a hand-written letter, in which case
the label is the actual letter represented by this image.
Other examples include DNA sequence identification,
recognition of an illness based on a set of symptoms,
speech recognition, and many others.

More formally, the objects are drawn independently from the object space $X$ (usually $X=[0,1]^d$ or
$\R^d$) according to some unknown but fixed probability distribution $P$ on $X$, and
labels are defined according to some function $\eta:X\rightarrow Y$, where
$Y$ is a finite set (often $Y=\Y$). The task is to construct a function
$\phi:\X\rightarrow Y$
which  approximates $\eta$, i.e. for which  $P\{x: \eta(x)\ne \phi(x)\}$ is small,
where $P$ and $\eta$ are unknown but examples $x_1,y_1,\dots,x_n,y_n$ are given;
 $y_i:=\eta(x_i)$. In the framework of statistical learning theory
 \cite{vc},\cite{vap} it is assumed that the function $\eta$ belongs
to some known class of functions $\C$.
Good error estimated can be obtained if the class $\C$ is small enough.
More formally, the number of examples required to obtain a
certain level of accuracy (or the \emph{sample complexity} of $\C$)
is linear in the VC-dimension of $\C$.

In this work we investigate the question whether such bounds can
be obtained if we consider only computable (on some Turing
machine) pattern recognition methods. To make the problem more
realistic, we also assume that the target function $\eta$ is also
computable. Both the predictors and the target functions are 
of the form $\{0,1\}^\infty\rightarrow\Y$.

We show that  there are classes $\C_k$ of
functions for which the  number of examples needed to approximate
the pattern recognition problem to a certain accuracy grows faster
in the VC dimension of the class than any computable function
(rather than being linear as in the statistical setting). In
particular this holds if $\C_k$ is the class of all computable
functions of length not greater than $k$.

Importantly, the same negative result holds even if we allow the
data to be  generated ``actively'', e.g. by some algorithm, rather
than just  by some fixed probability distribution.

To obtain this negative result we consider the task of data compression:
 an impossibility result for the task of data compression
allows us to estimate the sample complexity for pattern recognition.
  We also analyze how tight is the negative result, and show
that for some simple computable rule (based on the nearest neighbour
estimate) the sample complexity is finite in $k$,
under different definitions of  computational patterning recognition task.

In comparison to the vast literature on pattern recognition  and related
learning problems relatively little attention had been paid to the ``computable''
version of the task; at least this concerns the task of approximating
any computable function. 
There is a track of research in which different concepts of computable learnability 
of functions on countable domains are studied, see \cite{stl}. A link between this framework  
and  statistical learning theory is proposed in \cite{ms}, where it is argued that
for a uniform learnability finite VC dimension is required.

Another approach is 
to consider pattern recognition methods as  functions computable in polynomial time, or
under other resource constraints.
This approach leads to many interesting results, but it usually considers
more specified  settings of a learning problem, such as learning DNFs, finite automata, etc.
See  \cite{kea} for an introduction to this theory and for references.


\section{Preliminaries}\label{sec:pre}
A \emph{(binary) string} is a member of the set $\{0,1\}^*=\cup_{i=0}^\infty\{0,1\}^n$.
The length of a string
$x$ will be denoted by $|x|$, while $x^i$ is  the $i$th element
of $x$,  $1\le i\le|x|$. For a set $A$ the symbol $|A|$ is used for the number of elements in $A$.
We  will assume the lexicographical order on the set of strings, and when necessary
will identify $\X$ and $\N$ via this ordering.
Let  $\N$ be the sets of natural  numbers.
 The symbol $\log$ is used for $\log_2$. For
a real number $\alpha$ the symbol $\ulcorner\alpha\urcorner$ is
the least natural number not smaller than $\alpha$.

In pattern recognition a labelling function is usually a function from the interval $[0,1]$
or $[0,1]^d$ (sometimes more general spaces are considered) to a finite space $Y:=\{0,1\}$.
As we are interested in computable functions, we consider instead the functions
 of the form $\{0,1\}^\infty\rightarrow\Y$.
Moreover,
we call a partial recursive function (or program) $\eta$ a \emph{labelling function} if  there
exists such $t=:t(\eta)\in\N$ that $\eta$
accepts all strings from $X_t:=\{0,1\}^t$ and only such strings
\footnote{It is not essential for this definition that $\eta$ is not a total function.
An equivalent (for our purposes) definition would be as follows. A  labelling
function is any total function which outputs the string 00 on all inputs
except on the strings of some length $t=:t(\eta)$, on each of which it outputs either $0$ or $1$.}.
For an introduction to the computability theory see for example \cite{comp}.

It can be argued that this definition of a labelling function is too restrictive
to approximate well the notion of a real function. However, as we are after
negative results (for the class of all labelling functions),
it is not a disadvantage. Other possible definitions
are discussed in Section~\ref{sec:obr}, where we are concerned with tightness
of our negative results.

All computable function can be encoded (in a canonical way) and thus 
the set of computable functions can be effectively enumerated.
Define the \emph{length} of $\eta$ as $l(\eta):=|n|$ where $n$ is the minimal number of $\eta$ in such enumeration.

Define the task of computational pattern recognition as follows.
An (unknown) labelling function $\eta$ is fixed.
The objects $x_1,\dots,x_n\in X$
are drawn according to some distribution $P$  on $X_{t(\eta)}$.
The labels $y_i$ are defined according to  $\eta$, that
is $y_i:=\eta(x_i)$.

A \emph{predictor} is a family of functions (indexed by $n$)
$$
\phi_n(x_1,y_1,\dots,x_n,y_n,x),
$$ taking values in $Y$, such that
for  any $n$ and any $t\in\N$, if $x_i\in X_t$ for each $i$, $1\le i\le n$, then the
marginal $\phi(x)$ is  a total recursive function on  $X_t$ (that is,
$\phi_n(x)$ accepts any $x\in X_t$).
We will often identify $\phi_n$ with its marginal $\phi_n(x)$ when
the values of other variables are clear.

Thus, given a \emph{sample} $x_1,y_1,\dots,x_n,y_n$ of labelled objects of the same
size $t$, a predictor produces a computable function; this function is supposed
to  approximate the labelling function $\eta$ on $X_t$.

A \emph{computable predictor} is a predictor which for any $t\in\N$ and any $n\in\N$
 is a total recursive function on $X_t\times Y\times\dots\times X_t\times Y\times X_t$

\section{Main results}\label{sec:main}
We are interested in what size sample is required to approximate
a labelling function $\eta$.
Moreover, for a (computable) predictor $\phi$, a labelling function $\eta$ and $0<\epsilon\in\R$ define
\begin{multline*}
\delta_n(\phi,\eta,\epsilon):= \sup_{P_t} P_t\Big\{x_1,\dots,x_n\in X_t: \\
  P_t\big\{x\in X_t : \phi_n(x_1,y_1,\dots,x_n,y_n,x)\ne\eta(x)\big\}>\epsilon\Big\},
\end{multline*}
where $t=t(\eta)$ and $P_t$ ranges over all distributions on $X_t$. For $\delta\in\R$, $\delta>0$ define the
\emph{sample complexity} of $\eta$ with respect to $\phi$ as
$$
N(\phi,\eta,\delta,\epsilon):=\min\{n\in\N: \delta_n(\phi,\eta,\epsilon)\le\delta\}.
$$
The number $N(\phi,\eta,\delta,\epsilon)$ is the minimal sample size required
for a predictor $\phi$ to achieve $\epsilon$-accuracy with
probability $1-\delta$ when the (unknown) labelling function is $\eta$.

We can use statistical learning theory \cite{vc} to derive the following
statement
\begin{proposition} There exists a predictor $\phi$ such that
$$
N(\phi,\eta,\delta,\epsilon)\le \max\Big( l(\eta)\frac{8}{\epsilon}\log\frac{13}{\epsilon}, \frac{4}{\epsilon}\log\frac{2}{\delta}\Big)
$$
for any labelling
function $\eta$ and any $\epsilon,\delta>0$.
\end{proposition}
Observe that the bound is linear in the length of $\eta$.

In what follows the proof of this simple statement, we investigate the question of
whether any such bounds exist if we restrict our attention to computable predictors.

\begin{proof}
The predictor $\phi$ is defined as follows. For each sample $x_1,y_1,\dots,x_n,y_n$
it finds a shortest program $\bar\eta$ such that $\bar\eta(x_i)=y_i$ for all $i\le n$.
Clearly, $l(\bar\eta)\le l(\eta)$.
Observe that the VC-dimension of the class of all functions of length not greater
than $l(\eta)$ is bounded from above by $l(\eta)$, as there are not more than
$2^{l(\eta)}$ such functions. Moreover, $\phi$
minimises empirical risk over this class of functions.
It  remains to use the following bound (see e.g. \cite{dev}, Corollary~12.4)
$$
\sup_{\eta\in\mathcal C}N(\phi,\eta,\delta,\epsilon)\le \max\Big( V(\mathcal C)\frac{8}{\epsilon}\log\frac{13}{\epsilon}, \frac{4}{\epsilon}\log\frac{2}{\delta}\Big)
$$
where $V(\mathcal C)$ is the VC-dimension of the class $\mathcal C$.
\end{proof}

The main result of this work is that for any
computable predictor $\phi$ there is no computable upper bound in terms of $l(\eta)$
on the sample complexity of the function $\eta$ with respect to $\phi$:

\begin{theorem}\label{th:main} For any computable
predictor $\phi$  and any total recursive
function $\beta: \N\rightarrow\N$ there exist a labelling function $\eta$,
 and some $n>\beta(l(\eta))$ such that
$$
P\{x\in X_{t(\eta)}: \phi(x_1,y_1,\dots,x_n,y_n,x)\ne\eta(x)\}>\ep,
$$
 for any $x_1,\dots,x_n\in X_{t(\eta)}$,  where $y_i=\eta(x_i)$ and $P$ is the uniform
  distribution on $X_{t(\eta)}$.
\end{theorem}

For example, we can take  $\beta(n)=2^n$, or $2^{2^n}$.

\begin{corollary} For any computable predictor $\phi$, any total  recursive
function $\beta: \N\rightarrow\N$ and any $\delta<1$
$$
 \sup_{\eta: l(\eta)\le k}N(\phi,\eta,\delta,\ep)>\beta(k)
$$
from some $k$ on.
\end{corollary}

Observe  that there is no $\delta$ in the formulation of  Theorem~\ref{th:main}.
 Moreover, it is not important how the objects $(x_1,\dots,x_n)$ are generated~---
 it can be any individual sample.
In fact, we can assume that the sample is chosen in any manner, for example
by some algorithm. This means that no computable upper bound
on sample complexity exists even for \emph{active learning algorithms}.

It appears that the task of pattern recognition is closely related to
another learning task ---  data compression. Moreover,
to prove Theorem~\ref{th:main} we need a similar negative result
for this task.  Thus before proceeding with the proof of
the theorem, we introduce the task of data compression and derive some
negative results for it.
We call a  total recursive function $\psi:\X\rightarrow\X$ an \emph{data compressor} if it is an injection
(i.e. $x_1\ne x_2$ implies $\psi(x_1)\ne\psi(x_2)$).
We say that an data compressor \emph{compresses} the string $x$ if $|\psi(x)|<|x|$.
Clearly, for any natural $n$ any data compressor compresses not more than a half
of strings of size not greater than $n$.

We will now present a definition of Kolmogorov complexity; for fine
details see \cite{livi}, \cite{zl}.
The complexity of  a string $x\in\X$ with respect to a machine $\zeta$
is defined as
$$
 C_\zeta(x)=\min_p\{l(p):\zeta(p)=x\},
$$
where $p$ ranges over all partial functions (minimum over empty set is defined as $\infty$).
There exists such a machine $\zeta$ that $C_\zeta(x)\le C_{\zeta'}(x)+c_{\zeta'}$
for any $x$ and any machine $\zeta'$ (the constant $c_{\zeta'}$
depends on $\zeta'$ but not on $x$).
Fix any such $\zeta$
and define
the \emph{Kolmogorov complexity} of a string $x\in\X$ as
$$
C(x):=C_\zeta(x).
$$
Clearly, $C(x)\le |x|+b$ for any $x$ and for some $b$ depending only on $\zeta$.
A string is called $c$-incompressible if $C(x)\ge|x|-c$.
Obviously, any data compressor can not
compresses many $c$-incompressible strings, for any  $c$. However,
highly compressible strings (that is, strings with Kolmogorov complexity low
relatively to their length) might be expected to be compressed well by some sensible data compressor.
The following lemma shows that it can not be always the case, no matter
what we mean by ``relatively low''.

The proof of this lemma is followed by the proof of Theorem~\ref{th:main}.
\begin{lemma}\label{th:comp} For any data compressor $\psi$ and any total recursive function $\gamma:\N\rightarrow\N$
such that $\gamma$ goes monotonically to infinity
there exists a binary string $x$  such that $C(x)\le\gamma(|x|)$ and $|\psi(x)|\ge |x|$.
\end{lemma}
\begin{proof}
Suppose  the contrary, i.e. that  there exist an data compressor $\psi$ and some function $\gamma:\N\rightarrow\N$ monotonically
increasing to infinity
such that for any string $x$ if $C(x)\le\gamma(|x|)$ then $\psi(x)< |x|$.
Let  $T$ be the set of all strings which are not compressed by $\psi$
$$
T:=\{x:|\psi(x)|\ge|x|\}.
$$

Define the function $\tau$ on the set $T$ as follows: $\tau(x)$
is the number of the element $x$ in $T$
$$
\tau(x):=\#\{x' \in T:  x'\le x \}
$$
for each $x\in T$.
Obviously, the set $T$ is infinite.   Moreover,  $\tau(x)\le x$ for any $x\in T$
(recall that we identify $\X$ and $\N$ via lexicographical ordering).
Observe that $\tau$ is a total recursive function on $T$ and onto $\N$. Thus $\tau^{-1}:\N\rightarrow\X$ is
a total recursive function on $\N$. Thus,
 for any $x\in T$,
\begin{equation}\label{eq:raz}
 C(\tau(x))\ge C(\tau^{-1}(\tau(x))-c= C(x)-c>\gamma(|x|)-c,
\end{equation}
 for   constant $c$ depending only on $\tau$, where the first inequality follows from computability of $\tau^{-1}$ and the last
from the definition of $T$.

It is a well-known result (see e.g. \cite{livi}, Theorem~2.3.1) that
for any partial function $\delta$ that goes monotonically to infinity there is $x\in\X$
such that $C(x)\le\delta(|x|)$.
In particular, allowing $\delta(|x|)=\gamma(|x|)-2c$,
we conclude that there exist such $x\in T$ that
$$
 C(\tau(x))\le \gamma(|\tau(x)|)-2c\le \gamma(|x|)-2c,
$$
which contradicts~(\ref{eq:raz}).
\end{proof}

\emph{Proof of Theorem~\ref{th:main}.}
Suppose the contrary, that is that there exists such a computable predictor $\phi$
and a total
function $\beta: \N\rightarrow\N$ such that for any  labelling function $\eta$,
and any $n>\beta(l(\eta))$ we have
$$
P\{x: \phi(x_1,y_1,\dots,x_n,y_n,x)\ne\eta(x)\}\le\ep,
$$
 for some $x_i\in X_{t(\eta)}$, $y_i=\eta(x_i)$, $i\in\N$, where $P$ is the uniform distribution on $X_{t(\eta)}$.

Not restricting generality we can assume that $\beta$ is strictly increasing.
Define the (total) function $\beta^{-1}(n):=\max\{m\in\N :\beta(m)\le n\}$.
Define $\epsilon:=\ep$.
Construct the data compressor $\psi$ as follows. For each $ y\in\X$ define
$m:=|y|$, $t:=\ulcorner \log m\urcorner$.
Generate (lexicographically) first $m$ strings of length $t$ and denote
them by $x_i$, $1\le i\le m$. Define the labelling function $\eta_y$ as follows: $t(\eta_y)=t$ and
$\eta_y(x_i)=y^i$, $1\le i\le m$. Clearly, $C(\eta_y)\le C(y)+c$, where
$c$ is some universal constant capturing the above description.

Let $n:=\sqrt{m}$.
Next we run the predictor $\phi$ on all possible tuples
${\bf x}=(x_1,\dots,x_n)\in X_t^{\phantom{t}n}$ and each time count
 errors that $\phi$ makes on all elements of $X_t$:
$$
E({\bf x}):=\{x\in X_t: \phi(x_1,y^1,\dots,x_n,y^n,x)\ne \eta_y(x)\}.
$$
If $|E({\bf x})|>\epsilon m$ for each ${\bf x}\in X_t$
then $\psi( y):=0y$.

Otherwise proceed as follows. Fix some  tuple ${\bf x}=(x'_1,\dots,x'_n)$ such that $|E({\bf x})|\le\epsilon m$,
and let $H:=\{x'_1,\dots,x'_n\}$ be the unordered tuple $\bf x$.
Define
$$
\kappa^i:=\left\{ \begin{array}{ll}
e_0 & x_i\in E({\bf x})\backslash H, y^i=0\\
e_1 & x_i\in E({\bf x})\backslash H, y^i=1\\
c_0 & x_i\in  H, y^i=0\\
c_1 & x_i\in  H, y^i=1\\
* & \text{ otherwise }
\end{array}
\right.
$$
for $1\le i\le m$.
Thus,  each $\kappa^i$ is a member of a five-letter alphabet (a five-element set) $\{e_0,e_1,c_0,c_1,*\}$.
Denote the string $\kappa^1\dots\kappa^m$ by $K$.

Observe that the string $K$, the predictor $\phi$ and the order of $(x'_1,\dots,x'_n)$
(which is not contained in $K$) are sufficient to restore the string $y$.
Furthermore,
the $n$-tuple $(x'_1,\dots,x'_n)$ can be
obtained from $H$ (the un-ordered tuple) by the appropriate permutation;  let
$r$ be the number of this permutation in some fixed ordering of all $n!$ such
permutations. Using Stirling's formula, we have $|r|\le 2n\log n$; moreover,
to encode $r$ with some self-delimiting code we need not more than $4n\log n$
symbols (for $n>3$). Denote such encoding of $r$ by $\rho$.

Next, as there are $(1-\epsilon-\frac{1}{\sqrt{m}})m$ symbols $*$ in the $m$-element string $K$,
it can be encoded by some simple binary code $\sigma$ in such a way that
\begin{equation}\label{eq:count}
|\sigma(K)|\le \frac{1}{2}m + 7(\epsilon m +n).
\end{equation}
Indeed, construct $\sigma$ as follows. First replace all occurrences of the string $**$
with $0$.
Encode the rest of the symbols with any fixed 4-bit encoding such that the code of each letter
starts with $1$. Clearly, $\sigma(K)$ is uniquely decodable. Moreover, it is easy
to check that (\ref{eq:count}) is satisfied, as there are not less than $\frac{1}{2}(m-2(\epsilon m+n))$
occurrences of the string $**$. We also need to write $m$
in a self-delimiting way (denote it by $s$);
clearly, $|s|\le 2\log m$.

Finally, $\psi(\bar y)=1\rho s\sigma(K)$ and $|\psi( y)|\le |\bar y|$,
for  $m>2^{10}$.
Thus, $\psi$ compresses any $\bar y$ such that $n>\beta(C(\eta_y))$; i.e. such that
$\sqrt{m}>\beta(C(\eta_y))\ge\beta(C(y)+c)$.
This contradicts Lemma~\ref{th:comp} with $\gamma(k):=\beta^{-1}(\sqrt{k}-c)$.
\qed

\section{On tightness of the negative results}\label{sec:obr}
In this section we discuss how tight are the conditions of the
statements and to what extend they depend on the definitions.

Let us consider a question whether there exist any (not necessarily computable)
sample-complexity function
$$
 {\mathcal N}_\phi(k,\delta,\epsilon):=\sup_{\eta: l(\eta)\le k} N(\phi,\eta,\delta,\epsilon),
 $$
at least for some predictor $\phi$, or it is always infinity from some $k$ on.
\begin{proposition}\label{th:obr} There exist a predictor $\phi$ such that ${\mathcal N}_\phi(k,\delta,\epsilon)<\infty$
for any $\epsilon,\delta>0$ and any $k\in\N$.
\end{proposition}
\begin{proof} Clearly, $C(\eta)\ge C(t_{\eta})$.
Moreover,
$\liminf_{t\rightarrow\infty} C(t)=\infty$ so that
$$
  \max\{t_\eta:l(\eta)\le k\}<\infty
$$
for any $k$. It follows that the ``pointwise'' predictor
\begin{equation}
\phi(x_1,y_1,\dots,x_n,y_n,x)=\left\{ \begin{array}{ll}
y_i &\text{if } x=x_i, 1\le i\le n \\
0   & x\notin\{x_1,\dots,x_n\}
\end{array}
\right.
\end{equation}
satisfies the conditions of the proposition.
\end{proof}

It can be argued that probably this statement is due to our definition of a  labelling
function. Next we will discuss some other variants of this definition.

First, observe that if we define a labelling
function as any total function on $\X$  then some labelling
functions will not approximate any real function; for example such is
the function $\eta_+$ which counts bitwise sum of its input:
$
\eta_+(x):=\sum_{i=1}^{|x|}x_i \mod 2.
$
That is why we require  a labelling function to be defined only on $X_t$ for some $t$.

Another way to
 define a labelling function (which  perhaps makes labelling functions most close to real functions)
is as a function which accepts any \emph{infinite}  binary string.
Let us  call an \emph{i-labelling function}  any total recursive function
$\eta: \Y^\infty\rightarrow\Y$.  That is, $\eta$ is computable on a Turing
machine with an input tape on which one way infinite input is written, an
output tape and possibly some working tapes. The program $\eta$ is required
to halt on any input.
The next proposition shows that even if we consider such definition the situation
does not change. The definition of a labelling function $\eta$ in which
it accepts only finite strings is chosen in order to stay within conventional computability theory.
\begin{lemma}\label{th:inf} For any i-labelling function $\eta$ there exist
$n_\eta\in\N$ such that $\eta$ does not scan its input tape further position $n_\eta$.
In particular, $\eta(x)=\eta(x')$ as soon as $x_i=x'_i$ for any $i\le n_\eta$.
\end{lemma}
\begin{proof}
For any $x\in\X$ the program $\eta$ does not scan its tape further
some position $n(x)$ (otherwise $\eta$ does not halt on $x$). For
any $\chi\in\Y^\infty$ denote by $n_\eta(\chi)$ the maximal
$n\in\N$ such that $\eta$ scans the input tape up to the position
$n$ on the input $\chi$.

Suppose that $\sup_{\chi\in\Y^\infty}n_\eta(\chi)=\infty$, i.e.
that the proposition is false. Define $x^0$ to be the empty string. Furthermore,
let
$$
x^i=\left\{ \begin{array}{ll}
0   & \sup_{\chi\in\Y^\infty}n_\eta(x^1,\dots,x^{i-1}\chi)=\infty\\
1   &\text{ otherwise}
\end{array}
\right.
$$
By our assumption, $x_i$ is defined for each $i\in\N$.
Moreover, it easy to check that $\eta$ never stops on the input string $x_1x_2\dots$.
\end{proof}
Besides, it is easy to check that the number $n_\eta$ is computable.

Finally, it can be easily verified that
Proposition~\ref{th:obr} holds true if we consider i-labelling
functions instead of labelling functions, constructing the
required predictor based on the nearest neighbour predictor.
\begin{proposition} \label{th:obr2} There exist a predictor $\phi$ such that ${}^i{\mathcal N}_\phi(k,\delta,\epsilon)<\infty$
for any $\epsilon,\delta>0$ and any $k\in\N$, where ${}^i{\mathcal N}$ is defined as
${\mathcal N}$ with labelling functions replaced by i-labelling functions.
\end{proposition}
\begin{proof} Indeed, it suffices to replace the ``pointwise'' predictor in the proof of
Proposition~\ref{th:obr} by the
following predictor $\phi$, which assigns to the object $x$ the label
of that object among $x_1,\dots,x_n$ with whom $x$ has longest mutual prefix:
$
\phi(x_1,y_1,\dots,x_n,y_n,x):=y_k
$,
 where
 $$k:=\operatorname{argmax}_{1\le m\le n}\{\max\{i\in\N: x^1\dots x^i=x^1_m\dots x^i_m\}\};$$
to avoid infinite run in case of ties, $\phi$ considers only first (say) $n$ digits
of $x_i$ and break ties in favour of the lowest index.
\end{proof}

\end{document}